# A Method of Data Augmentation to Train a Small-Area Fingerprint Recognition Deep Neural Network with a Normal Fingerprint Database

Kim JuSong[1]

(1. Pyongyang Software Joint Development Center, Central District, Pyongyang, DPR of Korea)

**Abstract**

Fingerprints are popular among the biometric based systems due to ease of acquisition, uniqueness and availability. Nowadays it is used in smart phone security, digital payment and digital locker.

The traditional fingerprint matching methods based on minutiae are mainly applicable for large-area fingerprint and the accuracy rate would reduce significantly when dealing with small-area fingerprint from smart phone. There are many attempts to using deep learning for small-area fingerprint recognition, and there are many successes.

But training deep neural network needs a lot of datasets for training. There is no well-known dataset for small-area, so we have to make datasets ourselves.

In this paper, we propose a method of data augmentation to train a small-area fingerprint recognition deep neural network with a normal fingerprint database (such as FVC2002) and verify it via tests. The experimental results showed the efficiency of our method.

**Introduction**

Authentication and security are one of the major concerns of today's competitive and growing economy. Authentication based on passwords and PINs have a risk of being stolen or misused. In such a case, biometrics acts as an effective solution as it offers high security over these methods. Biometric traits can be either physiological like fingerprint, iris, face, palm or behavioral like gesture, gait etc. Out of these, fingerprints are the most commonly used biometric trait.

Currently, most mobile devices adopt very small fingerprint sensors that only capture small partial fingerprint images. Accordingly, conventional minutiae-based fingerprint matchers are not capable of providing convincing results due to the insufficiency of minutiae. To secure diverse mobile applications such as those requiring privacy protection and mobile payments, a more accurate fingerprint matcher is demanded. Although many approaches to fingerprint matching have been proposed, partial fingerprint matching for a small-area fingerprint scanner still remains a challenging task in fingerprint recognition, especially for mobile applications.

There are many attempts to using deep learning for small-area fingerprint recognition, and there many successes. To training deep neural network, we have to prepare large datasets. In particular, in the case of fingerprint image data, since it is personal information data and cannot be obtained sufficiently, the generalization performance of the trained neural network decreases as training proceeds with few data. In addition, as shown in the literature [2], if the images in the FVC2002 fingerprint database are cut into the required size to learn, the accuracy of the learning materials will be reduced and the learning will not proceed properly. And since fingerprint image data cannot be generated through simulation, the same method as in the literature [1] cannot be used.

From this, we propose one method to proceed with data augmentation from a small amount of general fingerprint image data, and verify the effectiveness of the proposed method through experiments by training neural networks from the data generated using this method.

**Extract Fingerprint Area**

Fingerprint databases such as FVC2002 have multiple images data for one fingerprint. Therefore, for every fingerprint, only one image is selected and processed. The fingerprint areas are calculated by the following method for the different fingerprint images selected in this way.

First, normalization was performed in the following way for the incoming image.

For grayscale fingerprint image $I$ with a size of $w \times h$, the normalized image $N$ was calculated as follows.

$$N(i,j) = \begin{cases} M_0 + \sqrt{\dfrac{V_0(I(i,j) - M)^2}{V}}, & I(i,j) > M \\ M_0 - \sqrt{\dfrac{V_0(I(i,j) - M)^2}{V}}, & I(i,j) \leq M \end{cases}$$

$$M = \frac{1}{w \times h} \sum_{i=0}^{w-1} \sum_{j=0}^{h-1} I(i,j)$$

$$V = \frac{1}{w \times h} \sum_{i=0}^{w-1} \sum_{j=0}^{h-1} (I(i,j) - M)^2$$

At this time, $M_0$ and $V_0$ were set to 100, respectively.

After that, binarization was performed on the obtained normalized image $N$. At this time, the threshold was set to 100.

The actual fingerprint area was calculated from the binarized image. Next, the area corresponding to the image size to be used in the deep neural network is obtained from the center of the fingerprint area.

This area is utilized to obtain the necessary image from the actual input image. The results of obtaining 128x128 image from the input image are shown below. (Figure 1.)

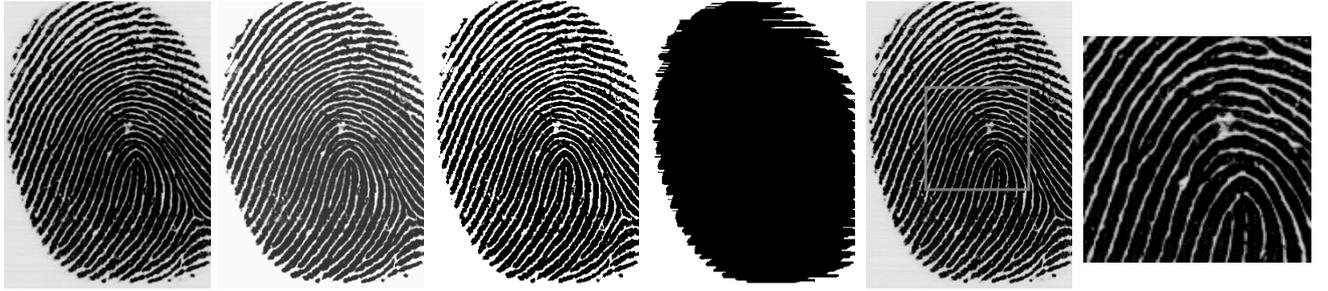

Figure 1. Process of obtaining 128x128 image from input image

### Data Augmentation

In general, when training deep neural networks that utilize image data, training proceeds with more data by augmenting the training data. Representative augmentation methods include rotation, shift, shear, zoom and flip, etc.

However, unlike general image classification or image identification, fingerprint recognition cannot utilize the general augmentation method. In the case of shearing, for example, the characteristics of the fingerprint (ridge orientation, minutiae…) may be changed, and the performance of the neural network will decrease when training with these image data. Zoom and flip cannot be used for the same reason.

So, we used rotation and shift.

As for rotation, first, an arbitrary value was selected in the 0˚ - 360˚, and used as a rotation angle. When the center point of the area obtained in the step above is $(i_s, j_s)$ and the size of the image to be obtained is $w_s \times h_s$, we extract the area of size $\left(2 \times \sqrt{\left(\frac{w_s}{2}\right)^2 + \left(\frac{h_s}{2}\right)^2}\right) \times \left(2 \times \sqrt{\left(\frac{w_s}{2}\right)^2 + \left(\frac{h_s}{2}\right)^2}\right)$ centered on $(i_s, j_s)$. Then, the obtained area is rotated by the selected rotation angle, and an image of size $w_s \times h_s$ is grown based on the center in the rotated area. The images thus obtained were added to the new database.

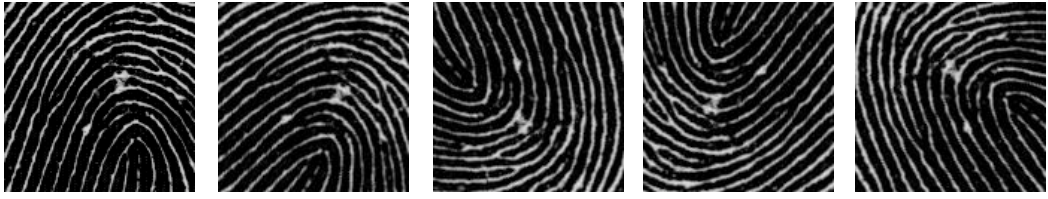

Figure 2. Augmented image by rotation (Normal image, 30°, 140°, 200°, 310°)

As for shift, the shift value was set as 80% or more overlaps the original image. That is, with respect to $(i_s, j_s)$, a value between $[i_s - (0.1 \times w_s), i_s + (0.1 \times w_s)]$ as the x-axis, and a value between $[j_s - (0.1 \times h_s), j_s + (0.1 \times h_s)]$ as the y-axis were selected and used. The images thus obtained were added to the new database.

Data augmentation can be carried out by rotation and shift, but it is not enough to increase generalization performance. The actual fingerprint image differs in brightness and contrast from the images in the database, and various noises are mixed. From this, in this paper, various augmentation methods were additionally used to improve generalization performance.

First, a method that can change the brightness and contrast of the image was used. Histogram Stretching and Histogram Equalization were used to change brightness and contrast.

In the histogram stretching, the brightness value $q$ of the output image from the brightness value $p$ of the input image was calculated as follows.

$$q = \frac{t_{max} - t_{min}}{p_{max} - p_{min}}(p - p_{min}) + t_{min}$$

Here, $p_{min}$ and $p_{max}$ are the minimum and maximum brightness values existing in the input image, and $t_{min}$ and $t_{max}$ are the maximum and minimum brightness values of the output image.

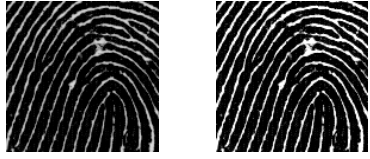

Figure 3. The results of the histogram stretching

Next, in histogram equalization, when the input brightness range is $[p_0, p_k]$ and the brightness range to be equalized is $[q_0, q_k]$, from the brightness value $p$, the brightness value of the output image $q$ is calculated as follows.

$$q = \frac{q_k - q_0}{w \times h} \sum_{i=p_0}^{p} H(i) + q_0$$

Here, $w$ and $h$ are the sizes of the input image, and $H(i)$ is a histogram value for the brightness value $i$.

In this way, the brightness and contrast were changed by using histogram stretching and histogram equalization.

To increase generalization performance, we added noise as the method for data augmentation. There are many ways to add noise, but Uniform Noise and Random Area Noise were added to simulate noise that may appear when collecting actual fingerprints.

Uniform Noise is a noise in which the brightness value of noise is uniformly distributed in a specific section, so it is suitable for images in which the brightness value is biased in a specific section, such as a fingerprint image.

$$p(z) = \begin{cases} \frac{1}{b-a}, & a \leq z \leq b \\ 0, & Else \end{cases}$$

Here, $z$ is the brightness value of the noise and $p(z)$ is the probability. And the distribution range of the brightness value of noise is $[a, b]$.

Next, in order to simulate noise that appears a lot when collecting fingerprints, random area noise was newly proposed and used. When collecting fingerprints, there are many phenomena in which some of the fingerprints are filled with black due to sweat. Therefore, in the paper, paint a certain part as black by selecting a region randomly.

First, the image was divided into a $w \times w$ sized window, and if the average brightness value $M$ is less than the threshold, it was determined as a dense fingerprint area, and fill a certain area with black from the center of the window. In the paper, the window size $w$ was set to 16, the threshold value was set to 64, and the average brightness value was calculated as follows.

$$M = \frac{1}{w \times w} \sum_{i=0}^{w-1} \sum_{j=0}^{w-1} I(i,j)$$

Next, the area to add noise is set by gradually moving a circle with a diameter of $w_n$ around the center point and overlapping four. The noise image with $w_n$ set to 32 below and four circles overlapped is shown.

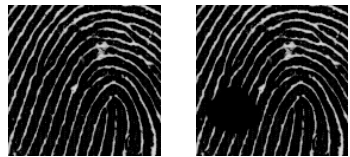

Figure 4. The fingerprint image with random area noise

### Setting the area of the extra image

Finally, an important step is to set the area in the extra images.

Fingerprint databases such as FVC2002 have several image data for one fingerprint. In the above steps, augmentation was performed by selecting one image for each fingerprint. For each fingerprint, the remaining images must be used to create a lot of learning data and improve generalization performance. However, if the area is set and augmented according to the above method for extra images, the accuracy of the dataset decreases. The reason is that even if the images on the fingerprint database are images of the same fingerprint, there are images that move and rotate a little, not the same position. Therefore, as in the literature [2], the method of cutting the center cannot be used, and if these images are not used, the generalization performance cannot be improved.

Because of this problem, in the paper, extract features and calculate similarity, and based on them, select the region. First, extract features for the reference area selected at the above step. And then, the remaining images for the same fingerprint were processed as follows.

The actual fingerprint area is obtained from the entire image according to the method performed in the < Extract Fingerprint Area> step. And then, the window is selected by the size of the input image used in the deep neural network. Features are extracted while moving this window sequentially with respect to the actual fingerprint area. The similarity is calculated by comparing the obtained features with the features of the reference area. If the largest value of the similarity calculated in this way is greater than a constant threshold, the area is set as the area to be used in the deep neural network and the image is cut off. Here, the feature extraction and similarity determination were carried out using the library of the fingerprint recognition engine that was already used in OEM modules.

The obtained area was augmented in the same way as in the <Data Augmentation> step.

In this way, the remaining images can also be used for learning to increase learning efficiency and generalization performance.

In this way, in the paper, a large amount of small-area fingerprint data with high accuracy were obtained from general fingerprint data and used to learning deep neural networks to improve recognition performance.

### Results and discussion

For performance evaluation, 2720 images for 415 fingerprints were prepared by synthesizing data from FVC2002DB1, Shengyuan, GA2016, and SY2018.

The file name was made into <Fingerprint Number>_<Identification Number>.bmp so that the same fingerprint could be recognized during learning.

Using the prepared images, 50 images were augmented for each image with individual augmentation methods raised in the paper to prepare learning dataset (Dataset 1). Next, by combining the augmentation methods raised in the paper, 30 images were augmented for each image to prepare learning dataset (Dataset 2). Next, as shown in Data2, 10 images were added with noise, and among them, 6 images with random area noise were prepared. (Dataset 3). Finally, the learning dataset (Dataset 4) were prepared by the method used in the literature [2].

The test was conducted using the Fingerprint_TF project (Network 1) and the deep neural network (Network 2) raised in the literature [2]. The size of the input image was set to 128x128.

Accuracy and Loss of test results are as follows.

| Dataset | Count of the train data | Count of the verify data | Network 1 | | Network 2 | |
|---|---|---|---|---|---|---|
| | | | Accuracy (%) | Loss (%) | Accuracy (%) | Loss (%) |
| **Dataset 1** | **135,000** | **1,000** | **99.1** | **0.9** | **98.3** | **1.7** |
| **Dataset 2** | **80,600** | **1,000** | **99.4** | **0.6** | **98.7** | **1.3** |
| **Dataset 3** | **80,600** | **1,000** | **99.5** | **0.5** | **98.8** | **1.2** |
| Dataset 4 | 2,220 | 500 | 95.2 | 4.8 | 93.4 | 6.6 |

Table 1. Results of deep neural network learning using augmented image dataset

In this way, it was confirmed that learning a deep neural network with small-area fingerprint recognition by augmenting general fingerprint image data improves performance.

In addition, through experiments, it was confirmed that it was effective to reuse various data augment methods (for example, to reuse histogram stretch while shifting and rotating as shown in Dataset 2).

Also, it was confirmed that it is recommended not to exceed 30% of the images added with noise, and among them, it is appropriate to make 60% of the images added with random area noise.